\pdfoutput=1

\documentclass[sigconf]{aamas}

\usepackage{balance} % for balancing columns on the final page
\usepackage{multirow}
\usepackage{tabularx}
\usepackage{array}
\usepackage{enumitem}
\usepackage{hyperref}
\usepackage{orcidlink}

\makeatletter
\gdef\@copyrightpermission{
  \begin{minipage}{0.2\columnwidth}
   \href{https://creativecommons.org/licenses/by/4.0/}{\includegraphics[width=0.90\textwidth]{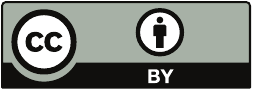}}
  \end{minipage}\hfill
  \begin{minipage}{0.8\columnwidth}
   \href{https://creativecommons.org/licenses/by/4.0/}{This work is licensed under a Creative Commons Attribution International 4.0 License.}
  \end{minipage}
  \vspace{5pt}
}
\makeatother

\setcopyright{ifaamas}
\acmConference[AAMAS '25]{Proc.\@ of the 24th International Conference
on Autonomous Agents and Multiagent Systems (AAMAS 2025)}{May 19 -- 23, 2025}
{Detroit, Michigan, USA}{Y.~Vorobeychik, S.~Das, A.~Nowé  (eds.)}
\copyrightyear{2025}
\acmYear{2025}
\acmDOI{}
\acmPrice{}
\acmISBN{}

\acmSubmissionID{Submission Number 167}

\title[Personality-Driven Decision-Making in LLMs]{Personality-Driven Decision-Making in LLM-Based Autonomous Agents}

\author{Lewis Newsham \orcidlink{0009-0007-3251-7716}}
\affiliation{
  \institution{Lancaster University}
  \city{Lancaster}
  \country{United Kingdom}}
\email{l.newsham1@lancaster.ac.uk}

\author{Daniel Prince \orcidlink{0000-0003-0962-8569}}
\affiliation{
  \institution{Lancaster University}
  \city{Lancaster}
  \country{United Kingdom}}
\email{d.prince@lancaster.ac.uk}

\begin{abstract}

    The embedding of Large Language Models (LLMs) into autonomous agents is a rapidly developing field which enables dynamic, configurable behaviours without the need for extensive domain-specific training. In our previous work, we introduced SANDMAN, a Deceptive Agent architecture leveraging the Five-Factor OCEAN personality model, demonstrating that personality induction significantly influences agent task planning. Building on these findings, this study presents a novel method for measuring and evaluating how induced personality traits affect task selection processes—specifically planning, scheduling, and decision-making—in LLM-based agents. Our results reveal distinct task-selection patterns aligned with induced OCEAN attributes, underscoring the feasibility of designing highly plausible Deceptive Agents for proactive cyber defense strategies.

\end{abstract}


\newcommand{\BibTeX}{\rm B\kern-.05em{\sc i\kern-.025em b}\kern-.08em\TeX}

\begin{document}

%%% The following commands remove the headers in your paper. For final 
%%% papers, these will be inserted during the pagination process.

\pagestyle{fancy}
\fancyhead{}

%%% The next command prints the information defined in the preamble.

\maketitle 

%%%%%%%%%%%%%%%%%%%%%%%%%%%%%%%%%%%%%%%%%%%%%%%%%%%%%%%%%%%%%%%%%%%%%%%%

\section{Introduction} \label{Introduction}

Autonomous agents are software or system entities that operate independently in an environment, capable of autonomous decision-making to achieve programmed objectives \cite{franklin1996agent, BOSSER20011002}. In the domain of cyber defense, \textit{Deceptive Agents} have emerged as a novel class of autonomous agent, designed to operate in decoy environments, and intended to deceive adversaries by replicating plausible human behaviours, thereby enabling an indistinguishable representation of a digital environment that is entirely fictitious in nature \cite{newsham2024sandman}. The aim of these agents thereby is to effectuate plausible mimesis for deception–a technique to signify the creation of a false belief \cite{pawlick2019game}. Therefore, Deceptive Agents are utilised to prolong attacker engagement and support intelligence gathering efforts whilst simultaneously deterring adversaries from production environments.

Recent efforts have explored using large-scale, pre-trained language models as the autonomous agent controller \cite{park2023generative, wang2023voyager, qian2023communicative}. This has resulted in a novel agent class referred to as \textit{Language Agents} \cite{wang2024survey, sumers2023cognitive}. A key rationale for using LLMs as the controller is that the agent is able to exploit the underlying LLM's extensive internal model of the world and its ability to capture long-range dependencies to support decision-making without domain-specific training \cite{vaswani2017attention}. Prior research in this area has demonstrated great potential in LLM-based agents completing often complex tasks which has subsequently led to the development of new frameworks and 'cognitive' architectures \cite{newsham2024sandman, sumers2023cognitive} which incorporate memory pipelines to aid long-term consistent decision-making. Notably, the role of the LLM is to provide both the decision-making and generative function in conjunction, whose outputs remain consistent with each other.

An important use of Language Agents is the representation of plausible human behaviour \cite{park2022social, park2023generative, newsham2024sandman}, often in collaborative and interactive environments \cite{yao2022react, nakano2021webgpt, wang2023voyager}. Such plausibility is crucial for the success of Deceptive Agents, which rely on realism to distract and deceive adversaries \cite{newsham2024sandman}. A key challenge, however, lies in crafting prompts that induce suitable personas in the LLM. While previous work has documented various prompt-engineering strategies for persona construction \cite{sumers2023cognitive, park2023generative, wang2023survey}, there is limited systematic exploration of how different persona prompts influence agent behaviour. One promising avenue is to leverage well-established psychological frameworks—such as the Five-Factor OCEAN model \cite{mccrae1992introduction, costa1999five}—as a foundation for persona prompt design.

Prior work on the SANDMAN architecture \cite{newsham2024sandman} demonstrated that prompt-based personality induction significantly affects schedule generation, showing how certain traits influence the creation and arrangement of tasks. Building on this foundation, the present study shifts focus from how schedules are generated to how they are subsequently employed. Specifically, we investigate how induced personality traits shape task selection and prioritisation in LLM-based agents—a dimension of autonomous decision-making that, to our knowledge, has not been previously examined. By transitioning from schedule generation to real-time decision-making with a pre-generated schedule, we highlight that induced traits continue to guide an agent’s behaviour well beyond the initial planning phase. Our core contributions are therefore:

\begin{itemize}[leftmargin=*, noitemsep]
    \item Proposing a method for measuring and evaluating the effect of prompt-based persona induction in context-dependent LLM-based agent decision-making;
    \item Evidence that prompt-based persona induction produces a stable yet non-deterministic effect on agent decisions that remain aligned with the induced persona trait.
\end{itemize}

The paper is structured as follows: Section \ref{Background} outlines background material and related work. Section \ref{Methodology} discusses the methodology, including our novel analytical approach. Section \ref{Results} details the results and discusses the findings. Lastly, Section \ref{Conclusion} concludes the paper.

% Introduction section (inc. Abstract) now fits perfectly on the first page
% Ideal for formatting and styling
% Provides room for acknowledgements at the end
\section{Background and Related Work} \label{Background}

Generative AI (GenAI) and LLMs have been used in various security applications to automate and streamline complex tasks, including software testing \cite{happe2023getting}, log parsing \cite{ma2024llmparser, setianto2021gpt}, and threat intelligence analysis \cite{bayer2023multi}. The extensive training of LLMs on internet-scale text endows them with emergent capabilities beyond generation and analysis. In many cases, LLMs appear to mimic or approximate forms of complex reasoning to achieve long-term objectives within interactive environments, often designed as multi-agent systems to enable collaboration \cite{yao2022react, nakano2021webgpt, wang2023voyager, qian2023communicative, park2023generative}. However, the exact nature of these `reasoning' capabilities–whether true reasoning or sophisticated pattern-matching–remains a subject of debate \cite{bubeck2023sparks}. Nonetheless, this capacity has given rise to a novel class of AI-enabled software agents known as \textit{LLM-based agents} \cite{guo2024large} or \textit{Language Agents} \cite{sumers2023cognitive}—systems that use LLMs as a core computational unit to reason, plan, and act.

Conventional autonomous agents excel at repetitive tasks in well-structured environments using heuristic policies or learned behaviours within defined constraints \cite{schulman2017proximal, mnih2015human, lillicrap2015continuous}. By contrast, Language Agents leverage the adaptability and expansive knowledge of LLMs, enabling natural interactions and a broader range of tasks. This flexibility stems from the transformer-based architecture \cite{vaswani2017attention}, combined with extensive training on large-scale text corpora, allowing them to perform diverse functions, including reasoning, planning, and dynamic interactions within their environment.

It is for this reason the SANDMAN Deceptive Agent framework \cite{newsham2024sandman} adopts an LLM-based agent approach to construct highly plausible simulacra of humans interacting with systems, acting as a honeypot with an enhanced degree of fidelity, depth, variance, and non-determinism to lure would-be attackers. A high level of plausibility is required to maintain the attacker's interest, enabling long-term intelligence gathering of the attacker's tools, tactics, and procedures (TTPs). In this context, Deceptive Agents are purposed to behave similarly to gray agents (computer-generated entities designed to simulate realistic, semi-adversarial participants) or non-player characters utilised in cyber-based exercises and autonomous cyber operations, such as the GHOSTS framework \cite{updyke2018ghosts}. Importantly, it is the LLM's capability to mimic human behaviour which makes them particularly useful for this novel form of automated deceptive behaviour.

Research has shown that LLMs can plausibly mimic human behaviours across various contexts, from cognitive tests and reasoning tasks \cite{dasgupta2022language, webb2023emergent, binz2023using, aher2023using, wong2023word} to complex simulations like social science experiments and micro-societies \cite{park2022social, park2023generative, ziems2024can}. Initial studies exploring the personalities of LLMs, inherently embedded or externally induced, have utilised psychometric tests such as the Big Five Inventory (BFI) \cite{john1991big} and IPIP-NEO \cite{goldberg1999broad, johnson2014measuring} to measure personality traits \cite{serapio2023personality, jiang2024evaluating}. The lexical hypothesis of personality, positing that significant personality traits are encoded in language, provides a theoretical foundation for studying how LLMs might embody human-like traits \cite{goldberg1981language, raad1998psycholexical, saucier2001lexical}.

These studies have demonstrated that LLMs, particularly pre-trained language models, implicitly encode aspects of human personality \cite{serapio2023personality}. Moreover, systematic prompt engineering has been shown to be an effective approach toward personality trait induction, leveraging the vast scale and pre-trained knowledge of LLMs without the need for model parameter adjustments \cite{jiang2024evaluating}. While these studies have successfully demonstrated that LLMs can exhibit synthetic personalities consistent with human psychological constructs, they have focused on measuring personality traits through psychometric inventories rather than exploring how induced personality traits affect behaviour in complex agent-based tasks \cite{jiang2024evaluating, pellert2023ai, serapio2023personality}.

In the context of Language Agent systems \cite{sumers2023cognitive}, the impact of induced personality on LLMs—particularly regarding autonomous planning and scheduling—remains underexplored. While agent-based implementations require consistent and predictable behaviours aligned with specific objectives \cite{wang2024survey}, few studies have evaluated how induced personality traits affect decision-making, task prioritisation, and overall agent performance in these frameworks.

Our previous work introduced `Deceptive Agents'—LLM-based systems designed to mimic human behaviour to deceive adversaries \cite{newsham2024sandman}. However, it primarily addressed forward task planning and scheduling without examining how induced personality traits influence autonomous decisions. Although recent studies have begun to simulate human personality traits in LLMs \cite{jiang2024evaluating, safdari2023personality}, empirical evidence on how these traits affect agent-based decision-making remains scarce. Understanding how personality induction shapes LLM-based agent behaviour is crucial not only for enhancing the plausibility of Deceptive Agents \cite{newsham2024sandman}, but also for enabling greater control over how task selection and prioritisation align with the induced personality in real-world scenarios. Accordingly, this study evaluates how induced personality traits influence decision-making in LLM-based agents, offering both empirical evidence of their impact and practical mechanisms for exerting control over these traits in real-world scenarios.
\section{Method} \label{Methodology}

In our prior work, a novel method to assess the impact of prompt-based persona induction using the OCEAN model on an LLM-based agent's forward task planning was proposed \cite{newsham2024sandman}. In this instance, the LLM-based agent was asked to construct a schedule of activites for a typical work-oriented day, based on a defined set of activities to choose from (\textit{e.g.}, taking a break, having lunch, writing a document, sending an email etc.). The initial, naive approach taken was then to execute these tasks as per the constructed schedule. However, this would not take into account the current context, or persona, of the Deceptive Agent when it became time to execute the activity from the schedule. This work therefore looks to expand upon this aspect, exploring the impact of prompt-based persona induction on activity selection, given the end-goal of each agent is to complete all the activities in a given schedule.

\subsection{Experimental Process} \label{Experimental Process}

To evaluate the effect of prompt-based persona induction on activity selection, a series of 500 work-oriented, daily schedules were generated by an LLM, each containing a range of activities which can be classified as either work (answering emails, writing a document) or personal (taking a break, having lunch), similar to activities used in previous research with LLM-based agents which adopt planning-based behaviours and simulate aspects of human behaviour \cite{park2023generative, qian2023communicative}. These 500 schedules were all varied in activity frequency, duration, ordering and type, and the schedule start time. The schedules were formatted in a machine-readable JSON format, with each activity being given a unique identifier (UID) to support systematic analysis. UIDs are generated based on a SHA-512 checksum of the task name and its assigned time. These machine-readable schedules then became the agent's `to-do' list with the agents' goal of undertaking all the activities on the to-do list.

In order to complete the to-do list, the language agent proceeds through a series of decision-cycles. At each decision-cycle, the agent is presented with the following zero-shot prompt format, comprising the components listed below in the specified order:

\begin{itemize}[leftmargin=*, noitemsep]
    \item \textbf{Personality:} A zero-shot expression of the intended persona, using either the positive or negative variant of the relevant personality trait from the OCEAN model, formatted according to the approach previously described in \citet{newsham2024sandman}.
    \item \textbf{Current Time:} A reference timestamp that grounds the decision-cycle chronologically, obtained by adding the duration of each completed task to the schedule’s initial start time.
    \item \textbf{Remaining To-Do List:} A JSON-based overview of all pending tasks, expressed in the format of: \texttt{Task Name (Duration), UID}.
    \item \textbf{Completed List:} A JSON-based record of tasks in the sequence they were chosen and finished, formatted as: \texttt{Task Name (UID)}.
    \item \textbf{Instructions:} Straightforward directives—such as “select the next task to perform” and “return only the task UID with no additional information”—urging the agent to consider only the contextual details provided.
\end{itemize}

\begin{figure}[htb]
    \centering
    \includegraphics[width=\linewidth]{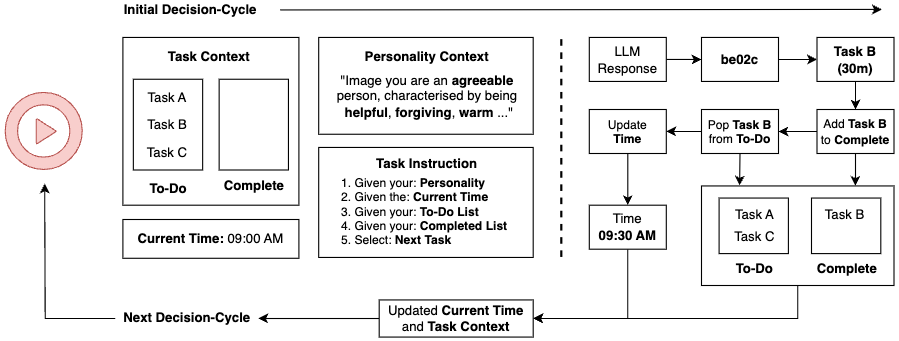}
    \caption{Decision-making task to be performed by the LLM featuring \underline{A}greeableness (Positive) as the induced trait.}
    \label{fig:Experiment}
    \Description{This image represents the decision-making task performed in this experiment. It is demonstrated at a high-level to depict the independent decision cycles the LLM is guided through as it processes one of the pre-generated schedules. The initial decision-cycle features a populated to-do list, and an empty completed list. The LLM is then provided personality context (induced with personality), and task instructions. The LLM then processes the schedule, providing a response in terms of which task it wishes to perform next, returning its unique task identifier (UID). This task UID is then removed from the to-do list, and added to the completed list. The current time is then updated based on this task's duration. This updated context is then passed back to the LLM within the next decision-cycle, where it is provided with the updated task context (to-do and completed lists).}
\end{figure}

The cyclical experimental process is shown in Figure \ref{fig:Experiment}. The persona statement is always supplied first to emphasise its importance in the LLM’s reasoning. The number of decision-cycles equals the total number of tasks in the to-do list; an experimental run concludes only once all tasks are completed, with no option to skip any. The sole experimental variable is the personality trait, which remains fixed for each run and across all 500 schedules, ensuring that any effects of induced personality are both isolated and measurable.

\subsection{Prompt-Based Persona Induction}

The prompt-based persona induction schema follows the one used in \citet{newsham2024sandman}. This uses the five-factor model of personality \cite{mccrae1992introduction, mccrae1997personality}, commonly known as OCEAN (\underline{O}penness, \underline{C}onscientiousness, \underline{E}xtraversion, \underline{A}greeableness, \underline{N}euroticism). The experiment induces personalities one trait at a time in one of two directions: forward (\textit{e.g.}, extraverted) or reverse (\textit{e.g.}, introverted). This method aligns with systematic approaches emplopyed in previous research on inducing personality traits in LLMs \cite{jiang2024evaluating, serapio2023personality, newsham2024sandman}.

The schema from our prior work \cite{newsham2024sandman} combines the naive descriptors and word-based characteristics identified in \cite{jiang2024evaluating} to construct personality trait statements. For instance, a personality trait of \underline{E}xtraversion, would be defined as follows:

\begin{itemize}[leftmargin=*, noitemsep]
\item \textbf{Naive}: "Imagine you are an \textbf{extraverted} person."
\item \textbf{Words}: "...characterised by being \textbf{outgoing}, \textbf{energetic}, \textbf{public}."
\item \textbf{Combined}: "Imagine you are an \textbf{extraverted} person, characterised by being \textbf{outgoing}, \textbf{energetic}, \textbf{public}."
\end{itemize}

This results in 10 experimental conditions, each representing one of the five personality traits (\underline{O}penness, \underline{C}onscientiousness, \underline{E}xtraversion, \underline{A}greeableness, \underline{N}euroticism) induced in a given direction (\textbf{Forward}:Positive, or \textbf{Reverse}:Negative), and is therefore consistent with prior research concerning the induction of personality traits in LLMs \cite{safdari2023personality, jiang2024evaluating, serapio2023personality}. The experiment also generates a control or `baseline' set of outputs used for comparison. The baseline outputs are generated in the same way but the LLM is not provided with a personality statement to consider.

The work thereby exploits the inherent non-determinism in LLMs, characterised by variability in outputs even with identical prompts \cite{ouyang2023llm}. Moreover, it also explores the impact of variation in the LLMs hyperparameters on the different induced personas. Specifically, we explore changes in the \textit{temperature} value.

\subsection{Analytical Approach}

A significant challenge of this experiment is how to measure the effect of the induced persona on task selection. To acheive this, the to-do list is conceived as a sequence defined by the order of the unique identifiers (UIDs) of its activities. The complete set of decision-cycles is considered a \textit{transformation process}. Finally, the completed list is treated as a newly generated transformed sequence defined by the new order of UIDs. This conceptualisation enables the analysis of two key properties of the transformation process. The first of concern is the \textit{plausibility} of the transformation, that is, to what extent is the transformation `behaviour' consistent with the induced personality trait. The second is determining the extent to which this activity selection process transforms the to-do list. By having measures of this transformation, it becomes possible to assess whether persona-based transformation is statistically significant from the baseline (no induced person) transformation. The analytical approaches here represent a novel contribution to approaches to systematically assess the impact of prompt engineering.

To assess plausibility, we use measurement of \textit{movement deltas}, which quantify the shifts in activity positions before and after the persona-based transformation. The UID of the activity is used to index and calculate the movement delta based on the initial and the resultant transformed position. Negative movement deltas indicates an activity is moved earlier and therefore considerered to be prioritised, and a positive movement delta indicates an activity is moved later and therefore is deprioritised. The traits of \underline{C}onscientiousness and \underline{E}xtraversion serve as the central focus in this analysis, which are hypothesised to affect task selections related to work efficiency and social interaction, respectively \cite{wilmot2019century, hogan1997conscientiousness, lucas2000cross}. 

%\subsubsection{Measures of Transformation} \label{Statistical Analysis}
Given the to-do and completed lists are treated as sequences defined by their UIDs, it is possible to apply standard approaches of sequence comparison for similarity and difference. Prior to the experimentation, it was not clear which standard measure of sequence similarity would yield the greatest analytical worth. Therefore, the following approaches were used to to produce quantitative assessments of the sequence transformation:

\begin{itemize}[leftmargin=*, noitemsep]
    \item \textbf{Longest Common Substring (LCSS)}: Finds the longest sequence of consecutive elements common to both sequences. Assesses impact on maintaining uninterrupted task sequences, focusing on contiguous matches.
    \item \textbf{Longest Common Prefix (LCP)}: Finds the longest initial segment common to both sequences. Reflects initial decision-making patterns by comparing the start of both sequences.
    \item \textbf{Levenshtein Distance}: Minimum number of single-character edits (insertions, deletions, or substitutions) required to transform one sequence into another. Provides a comprehensive view of all changes needed \cite{yujian2007normalized}.
    \item \textbf{Longest Common Subsequence (LCS)}: Measures similarity by finding the longest subsequence common to both strings. Indicates order preservation in task selection by identifying deletions and additions \cite{paterson1994longest}.
    \item \textbf{Similarity Ratio (SR)}: Normalised measure derived from LCS length. Evaluates similarity between to-do and completed lists by considering deletions and additions.
    \item \textbf{Hamming Distance}: Number of positions at which the corresponding elements differ. Applicable only for sequences of the same length and focuses on substitutions.
\end{itemize}

For each directional personality trait, 500 measurements of each metric were produced enabling a statistical significance test of the impact of the persona induction against the baseline.
\section{Results} \label{Results}

In this section, we present the outcomes of our analyses, which are divided into two key parts corresponding to the analytical approaches outlined earlier. Firstly, we explore the plausibility of the observed movement delta shifts in task prioritisation by the LLM after the induction of specific personality traits. This analysis focuses on evaluating whether the LLM's behaviour aligns with established psychological understandings of \underline{C}onscientiousness and \underline{E}xtraversion, particularly in relation to work-oriented and socially-related tasks, respectively. The goal here is to assess whether the LLM prioritises tasks in a manner consistent with the traits it has been induced with, thereby providing insights into the model's ability to mimic human-like decision-making patterns. Subsequently, we turn our attention to a comprehensive statistical analysis that examines the broader impact of inducing personality traits across all five dimensions of the OCEAN model \cite{mccrae1992introduction, costa1999five}, adopting a similar approach to previous research on personality traits in LLMs \cite{jiang2024evaluating, safdari2023personality, serapio2023personality, newsham2024sandman}. Using a suite of quantitative measures specifically designed to assess transformations in the overall ordering of tasks, we evaluate how the induced personality traits influence the LLM's task selection behaviour compared to a baseline condition, where no specific traits were introduced. This approach provides a robust framework for measuring changes in the sequential structure of the schedules, allowing us to quantify the consistency and reliability of personality-driven modifications in LLMs.

\subsection{Plausibility Analysis}

As discussed, to evaluate the impact regarding the plausibility of induced personality traits on the target LLMs, the \textit{movement delta shifts} of individual tasks is analysed. As previously defined, \textit{movement delta shifts} are the direction and magnitude of the change in task order from the initial to-do list to the completed list after the LLM processes and transforms the task execution sequence. These shifts are examined to determine if they align with the expected priorities of the induced traits. For example, an LLM with high \underline{C}onscientiousness (\textbf{CON-POS}) is expected, given the personality schema and method of induction, to prioritise tasks requiring organisation, discipline, and hard work, such as work-related activities \cite{wilmot2019century}. Movement delta shifts are quantified by calculating the mean shift ($\mu$) for each task, where negative $\mu$ values indicate prioritisation (the task appears earlier in the completed list), and positive $\mu$ values suggest deprioritisation.

In this analysis, we hypothesise that an LLM induced with high \underline{C}onscientiousness will prioritise work-centric tasks, while one induced with high \underline{E}xtraversion will favour socially engaging activities. In our analysis, tasks relevant to these traits are highlighted in yellow, while others are shown in blue. Given the advanced reasoning capabilities of the GPT-4o model, our analysis is restricted to this model, aiming to determine the plausibility of the LLM’s decision-making behaviour in aligning with expected task prioritisation patterns.

\subsubsection{\underline{C}onscientiousness: Work-related Activities} \label{Conscientiousness (Work Activities)}

Work-related tasks (highlighted in yellow) made available to the LLM include: Email, Planning, Work, Team Collaboration, Meeting, Research. Figure \ref{fig:Movement_Deltas_Positive_Conscientiousness} below illustrates the movement delta shift for all tasks following \textbf{positive} induction of \underline{C}onscientiousness.

\begin{figure}[htb]
    \centering
    \includegraphics[width=\linewidth]{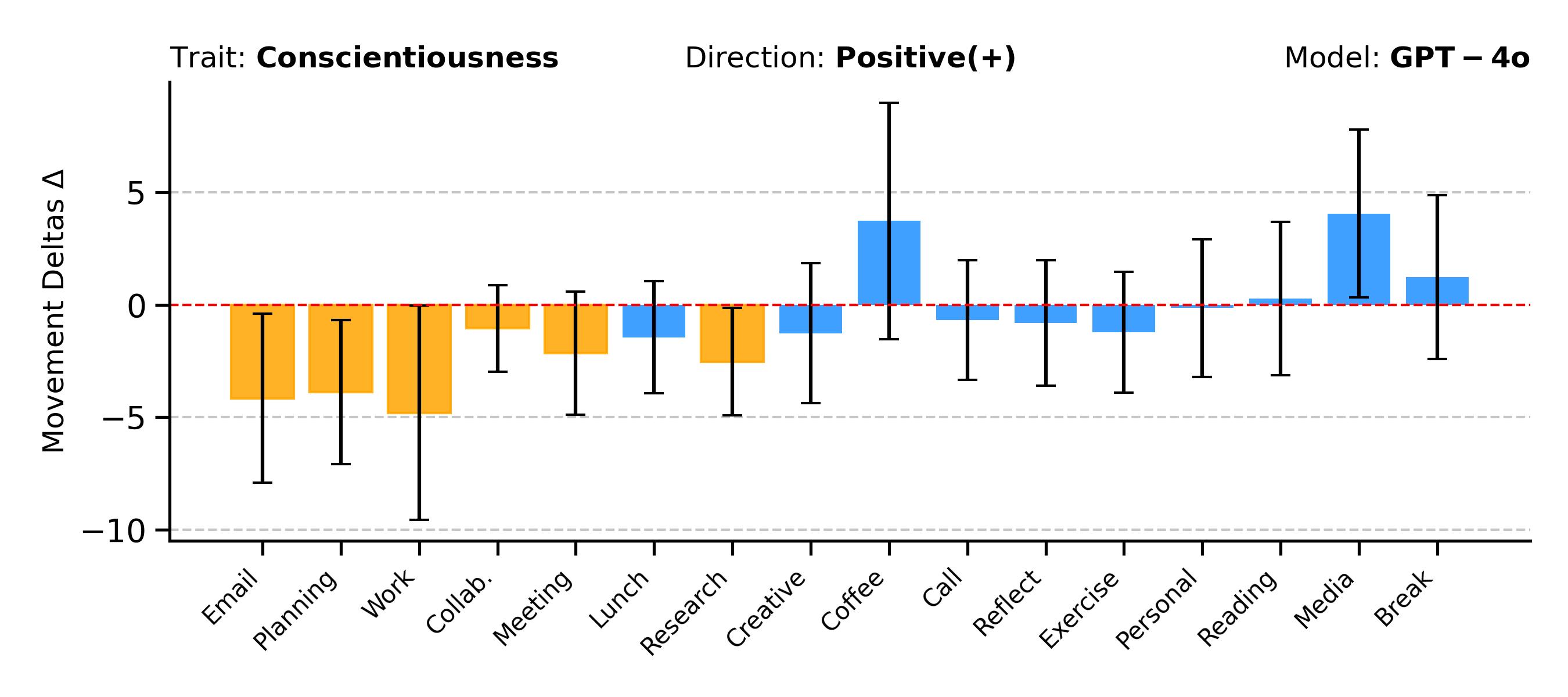}
    \captionsetup{font=small}
    \caption{Movement Deltas: Positive \underline{C}onscientiousness (GPT-4o).}
    \label{fig:Movement_Deltas_Positive_Conscientiousness}
    \Description{This image illustrates the movement delta shift of all tasks for the positively-induced trait of Conscientiousness. Notable shifts in the forward direction are observed for work-related tasks, namely Work, Email, Planning, Meeting, Research. Inversely, non-work-related tasks such as Coffee and Social Media are deprioritised.}
\end{figure}

The degree of \textit{plausibility} is inferred by comparing the LLM's task prioritisation to established psychological understandings of the specific trait induced. To justify our focus on the trait of \underline{C}onscientiousness in the context of work-related tasks, it is a trait identified as a key non-cognitive predictor of occupational performance and encompasses traits such as diligence, responsibility, and self-control \cite{wilmot2019century, hogan1997conscientiousness, roberts2014conscientiousness}. Therefore, following a positively-reinforced induction of this trait, it is logical to expect prioritisation of work-related tasks. This predicted effect is clearly demonstrated in Figure \ref{fig:Movement_Deltas_Positive_Conscientiousness} where the LLM (GPT-4o) clearly favours work-related tasks over others which may be considered non-work-related, such as Call, Reflective Time, and Exercise.

\begin{figure}[htb]
    \centering
    \includegraphics[width=\linewidth]{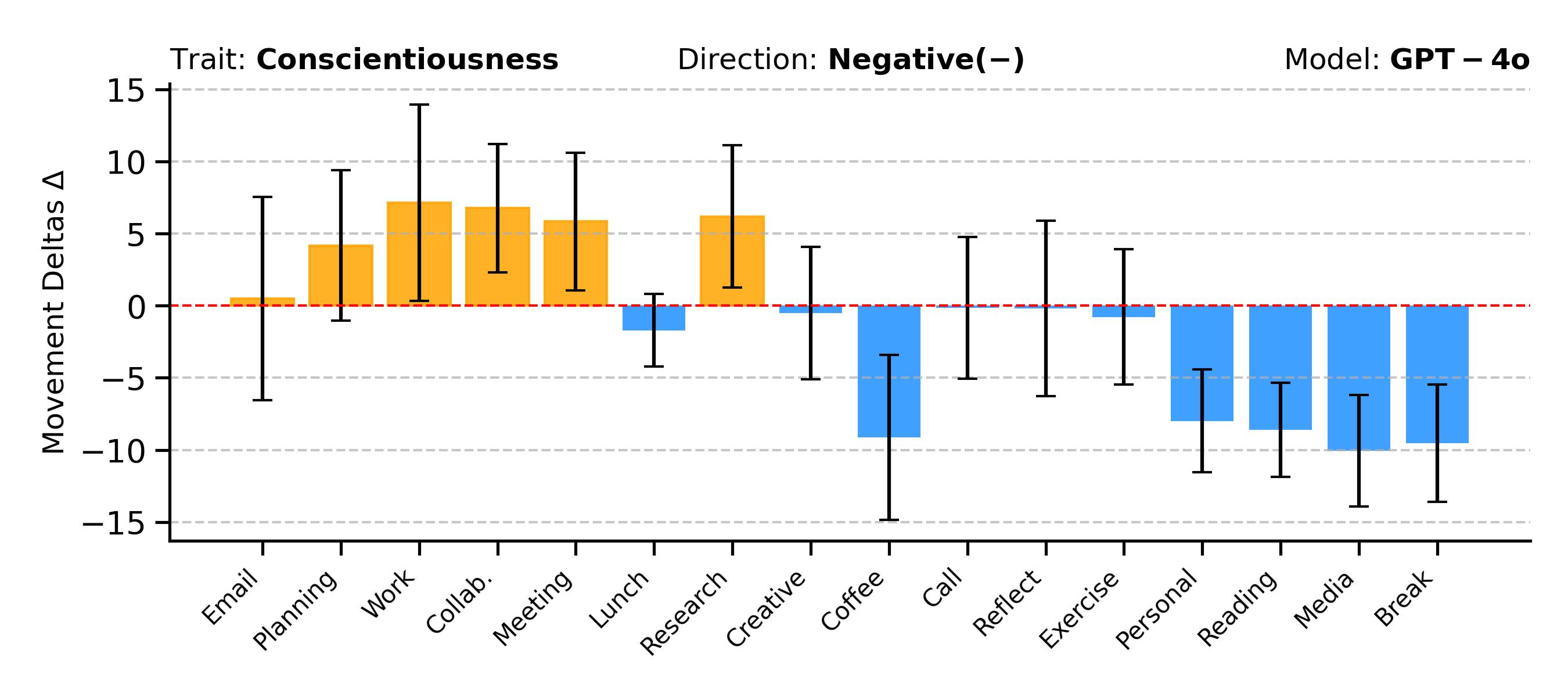}
    \captionsetup{font=small}
    \caption{Movement Deltas: Negative \underline{C}onscientiousness (GPT-4o).}
    \label{fig:Movement_Deltas_Negative_Conscientiousness}
    \Description{This image illustrates the movement delta shift of all tasks for the negatively-induced trait of Conscientiousness. Notable shifts in the forward direction are observed for non-work tasks such as Coffee Break, Personal Time, Social Media etc. whereas an inverse effect is also recorded in the reverse direction regarding work-related tasks, namely Work, Team Collaboration, Meeting etc.}
\end{figure}

In contrast, an individual low in \underline{C}onscientiousness may display disorganisation, a lack of discipline, and general disregard for work-related responsibilities \cite{wilmot2019century, hogan1997conscientiousness, robertson2000conscientiousness}. As shown in Figure \ref{fig:Movement_Deltas_Negative_Conscientiousness}, the LLM deprioritises work-related tasks in favour of non-work-related activities when negatively induced, such as Personal Time, Reading, and Social Media, demonstrating a significant shift in behaviour.

\subsubsection{Movement Delta Shift: Work}

\begin{figure}[htb]
    \centering
    \includegraphics[width=\linewidth]{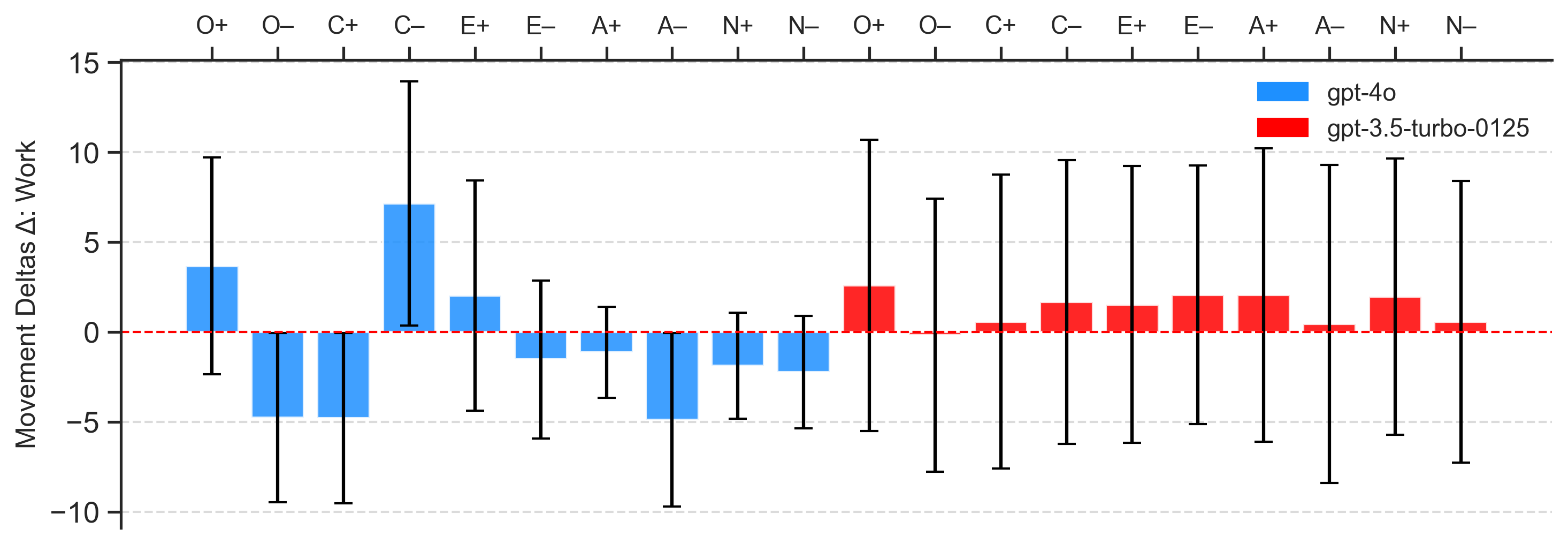}
    \captionsetup{font=small}
    \caption{Movement Deltas: \textbf{Work} (GPT-4o \& GPT-3.5-Turbo)}
    \label{fig:Movement_Deltas_Work}
    \Description{This image represents the movement delta shift of the Work task on all experimental conditions within both models. Each shift for the GPT-4o model is represented as blue bars, whereas GPT-3.5-Turbo is represented as red bars. Variance of each result, namely its standard deviation, is indicated by the error bar on each bar plot. Results for the GPT-4o model indicate significant deviations regarding the prioritisation of Work across all traits on both polarities. In constrast, the GPT-3.5-Turbo appears to deprioritise the task of Work across all traits to a similar extent (ranging from 0 to +3).}
\end{figure}

To provide a more comprehensive and task-specific analysis, we examine the movement deltas for the \textit{Work} task across all experimental conditions, as shown in Figure \ref{fig:Movement_Deltas_Work}. This figure compares the shifts observed for the GPT-4o model (blue bars) with those for the GPT-3.5-Turbo model (red bars), with error bars indicating the standard deviations.

Figure \ref{fig:Movement_Deltas_Work}, above, highlights a stark contrast in behaviour between GPT-4o and GPT-3.5-Turbo. GPT-4o shows greater alignment with expected behaviour based on the induced traits, particularly \underline{C}onscientiousness, exhibiting more dynamic movement delta shifts, whereas GPT-3.5-Turbo appears more deterministic, with minimal notable effects across the conditions.

\subsubsection{\underline{E}xtraversion: Social Activities} \label{Extraversion (Social Activities)}

\begin{figure}[htb]
    \centering
    \includegraphics[width=\linewidth]{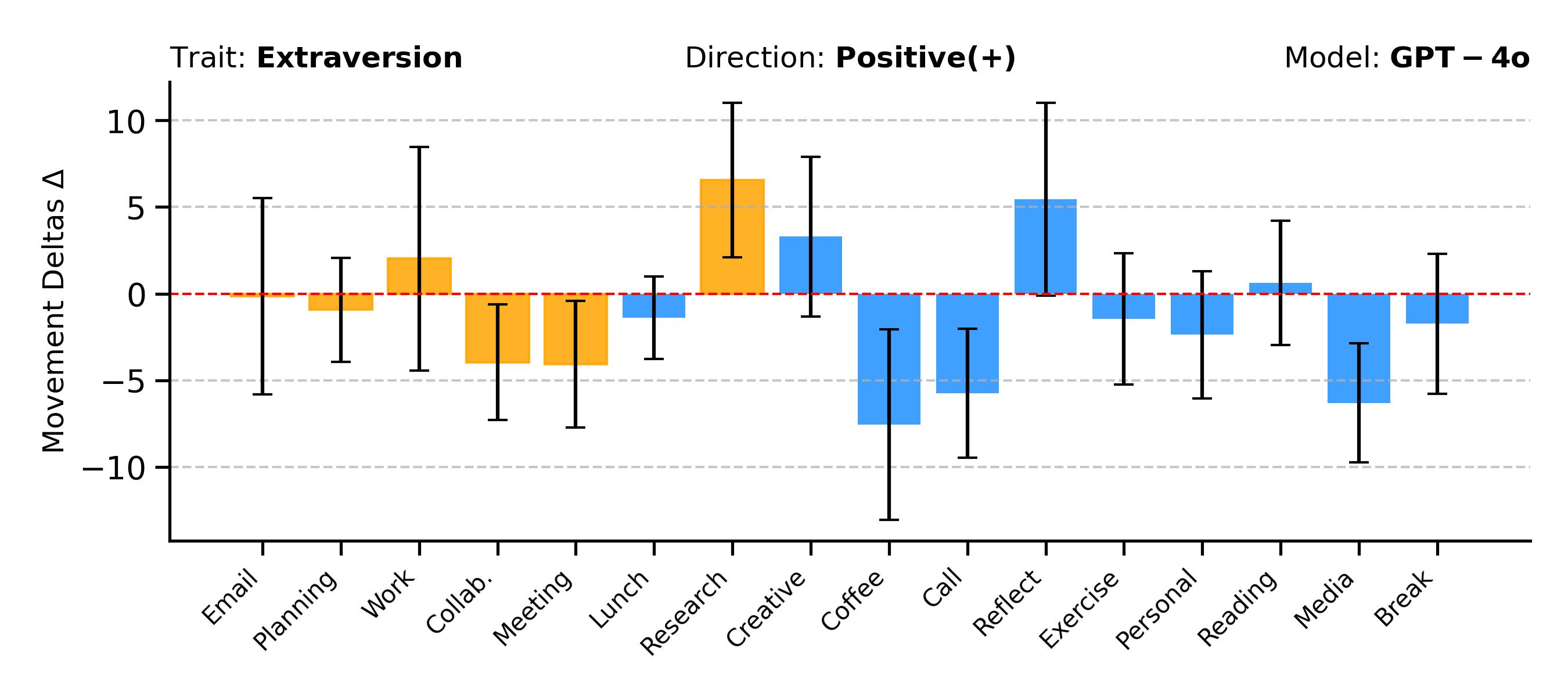}
    \captionsetup{font=small}
    \caption{Movement Deltas: Positive \underline{E}xtraversion (GPT-4o).}
    \label{fig:Movement_Deltas_Positive_Extraversion}
    \Description{This image illustrates the movement delta shift of all tasks for the positively-induced trait of Extraversion. Notable shifts in the forward direction are observed for social-related tasks such as Team Collaboration, Meeting, Call, Social Media. Inversely, tasks which may often not feature social interaction are observed in the reverse direction. For instance, the tasks of Research and Reflection are deprioritised.}
\end{figure}

Socially-related tasks (highlighted in yellow) made available to the LLM include: Team Collaboration, Meeting, Call, and Social Media. Figure \ref{fig:Movement_Deltas_Positive_Extraversion} represents the movement delta shift following positive induction of \underline{E}xtraversion.

As expected, the LLM prioritises social tasks, reflecting the core aspects of \underline{E}xtraversion, such as sociability and reward sensitivity \cite{lucas2000cross}. This is evident in the prioritisation of tasks like Team Collaboration, Meeting, and Call, along with other related activities such as Coffee Break, with slight increases in Exercise and Personal Time.

\begin{figure}[htb]
    \centering
    \includegraphics[width=\linewidth]{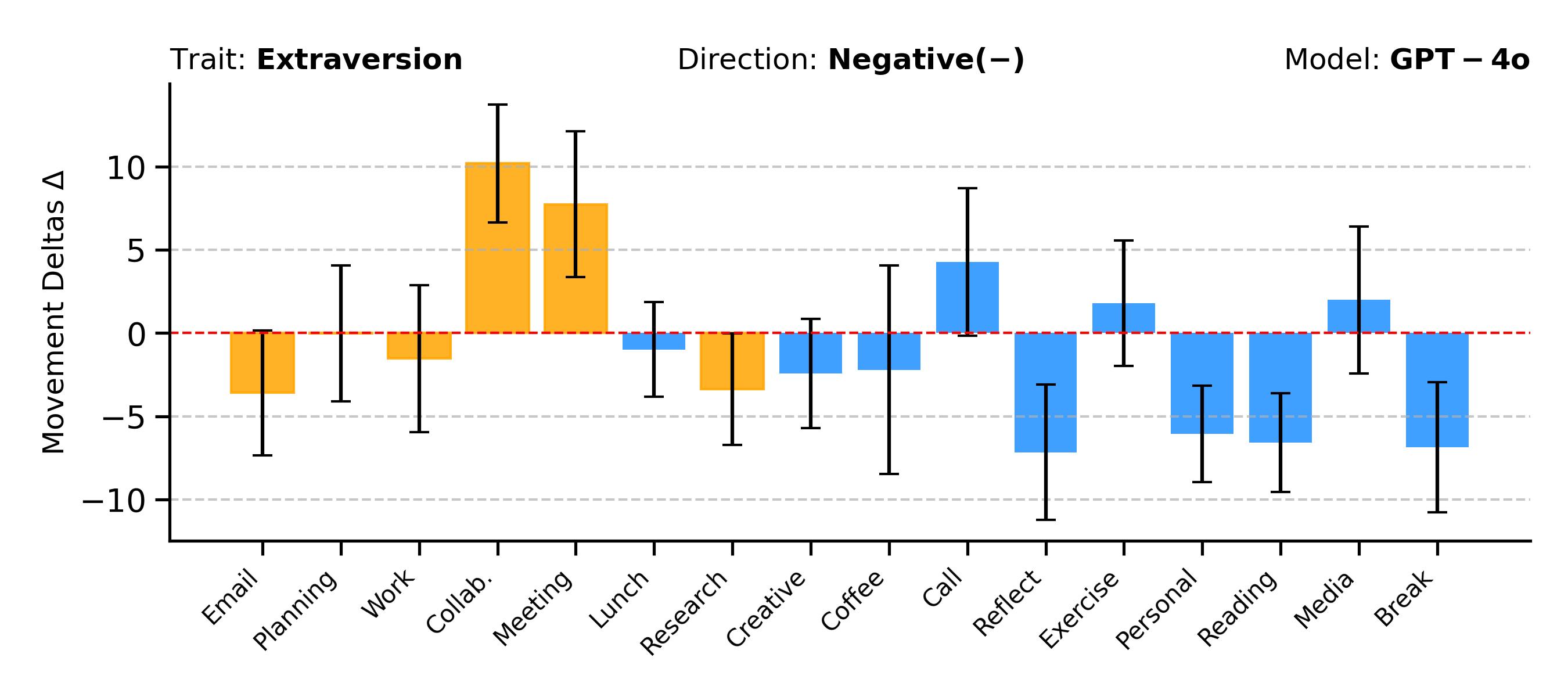}
    \captionsetup{font=small}
    \caption{Movement Deltas: Negative \underline{E}xtraversion (GPT-4o).}
    \label{fig:Movement_Deltas_Negative_Extraversion}
    \Description{This image illustrates the movement delta shift of all tasks for the negatively-induced trait of Extraversion. Notable shifts in the reverse direction are observed for social-related tasks such as Team Collaboration, Meeting, and Call. Inversely, a forward direction is observed for less socially oriented tasks, such as Research, Reflection, Personal Time, Reading, Break.}
\end{figure}

Conversely, as shown in Figure \ref{fig:Movement_Deltas_Negative_Extraversion}, when negatively-reinforcing the trait of \underline{E}xtraversion, the LLM deprioritises socially engaging tasks, instead favouring more solitary activities such as Reflective Time, Personal Time, and Reading. This shift suggests that the LLM adopts introverted tendencies, prioritising tasks that are less socially oriented.

In summary, these results suggest that the method of trait induction used in this experiment effectively steers the LLM's behaviour in a manner consistent with the induced personality traits. The LLM's prioritisation of tasks is driven by the personality it has been induced with, without direct instruction to prioritise tasks relevant to that personality. For instance, after positive induction with \underline{E}xtraversion, the LLM naturally favors social-related tasks, leading to their earlier appearance in the completed list, as indicated by negative mean ($\mu$) movement delta shift values.

\begin{figure}[htb]
    \centering
    \includegraphics[width=\linewidth]{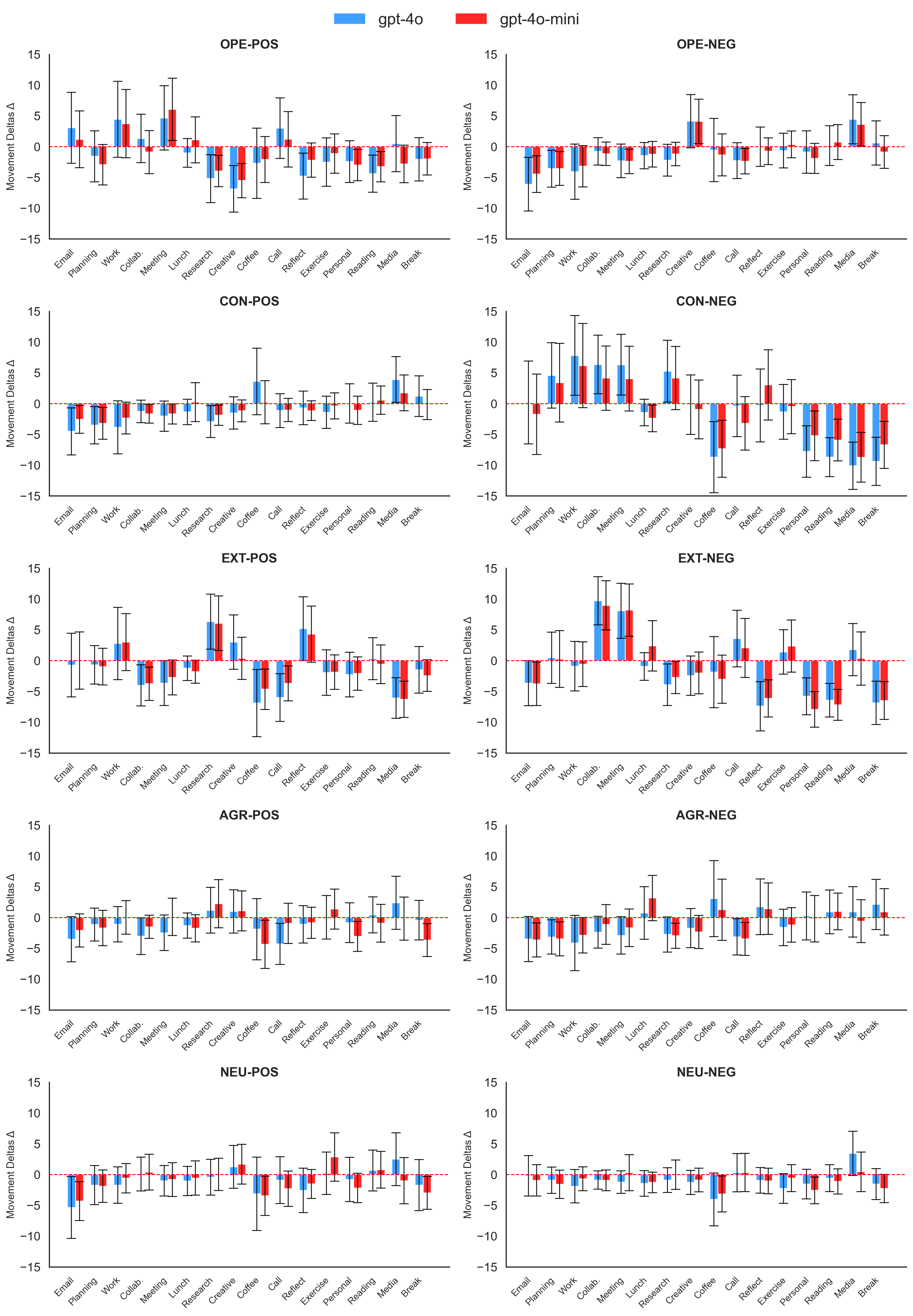}
    \captionsetup{font=small}
    \caption{Aggregated movement deltas per experimental condition. X-axis: Tasks. Y-axis: Average ($\mu$) movement delta shift, ranging from -15 to +15. Error bars denote deviation ($\sigma$).}
    \label{fig:Movement_Deltas_Tasks_Combined}
    \Description{This image presents the results across all OCEAN traits, on both polarities, using the models of GPT-4o and GPT-4o-Mini, on all available tasks. The results demonstrate both models exhibit similar behaviour in terms of task prioritisation, which is expected given they belong to the same model family but differ in terms of size and resource efficiency.}
\end{figure}

\begin{table*}[ht]
\setlength{\tabcolsep}{7.5pt}
\centering
\small

\caption{Normalised GPT-4o and GPT-4o-Mini results with adjusted sampling temperatures on neutral personality experiment. Temperature ($\tau$) range: \texttt{0.0} to \texttt{1.6}. Values are aggregated on the entire sample of pre-generated schedules ($n$ = 500).}

\label{tab:4o-4o-mini-temperature}
\begin{tabular}{@{}c c c c c c | c c c c c c c@{}}
\toprule
\multirow{2}{*}{\centering \textit{$\tau$}} 
& \multicolumn{5}{c}{\textbf{GPT-4o} ($\mu$ / $\sigma$)} 
& \multirow{2}{*}{\centering \textit{$\tau$}}
& \multicolumn{5}{c}{\textbf{GPT-4o-Mini} ($\mu$ / $\sigma$)} \\
\cmidrule(r){2-6} \cmidrule(r){8-12}
& \multicolumn{1}{c}{\textbf{LCSS}} & \multicolumn{1}{c}{\textbf{LCP}} & \multicolumn{1}{c}{\textbf{LEV}} & \multicolumn{1}{c}{\textbf{SR}} & \multicolumn{1}{c}{\textbf{HAM}} 
& & \multicolumn{1}{c}{\textbf{LCSS}} & \multicolumn{1}{c}{\textbf{LCP}} & \multicolumn{1}{c}{\textbf{LEV}} & \multicolumn{1}{c}{\textbf{SR}} & \multicolumn{1}{c}{\textbf{HAM}} \\
\midrule

0.0 & \textbf{0.635}\scriptsize{0.224} & \textbf{0.604}\scriptsize{0.260} & \textbf{0.201}\scriptsize{0.144} & \textbf{0.321}\scriptsize{0.220} & \textbf{0.873}\scriptsize{0.095} & 0.0 & \textbf{0.522}\scriptsize{0.249} & \textbf{0.476}\scriptsize{0.288} & \textbf{0.296}\scriptsize{0.178} & \textbf{0.393}\scriptsize{0.245} & \textbf{0.829}\scriptsize{0.111} \\

0.2 & \textbf{0.633}\scriptsize{0.245} & \textbf{0.599}\scriptsize{0.268} & \textbf{0.207}\scriptsize{0.148} & \textbf{0.313}\scriptsize{0.224} & \textbf{0.875}\scriptsize{0.094} & 0.2 & \textbf{0.506}\scriptsize{0.252} & \textbf{0.473}\scriptsize{0.285} & \textbf{0.305}\scriptsize{0.181} & \textbf{0.405}\scriptsize{0.244} & \textbf{0.824}\scriptsize{0.116} \\

0.4 & \textbf{0.624}\scriptsize{0.242} & \textbf{0.593}\scriptsize{0.274} & \textbf{0.209}\scriptsize{0.147} & \textbf{0.310}\scriptsize{0.228} & \textbf{0.879}\scriptsize{0.093} & 0.4 & \textbf{0.515}\scriptsize{0.250} & \textbf{0.473}\scriptsize{0.289} & \textbf{0.287}\scriptsize{0.175} & \textbf{0.384}\scriptsize{0.229} & \textbf{0.836}\scriptsize{0.111} \\

0.6 & \textbf{0.588}\scriptsize{0.222} & \textbf{0.561}\scriptsize{0.252} & \textbf{0.232}\scriptsize{0.150} & \textbf{0.328}\scriptsize{0.213} & \textbf{0.866}\scriptsize{0.097} & 0.6 & \textbf{0.501}\scriptsize{0.262} & \textbf{0.458}\scriptsize{0.293} & \textbf{0.317}\scriptsize{0.202} & \textbf{0.408}\scriptsize{0.248} & \textbf{0.819}\scriptsize{0.127} \\

0.8 & \textbf{0.578}\scriptsize{0.237} & \textbf{0.558}\scriptsize{0.258} & \textbf{0.235}\scriptsize{0.154} & \textbf{0.339}\scriptsize{0.229} & \textbf{0.857}\scriptsize{0.103} & 0.8 & \textbf{0.494}\scriptsize{0.242} & \textbf{0.450}\scriptsize{0.283} & \textbf{0.320}\scriptsize{0.188} & \textbf{0.429}\scriptsize{0.249} & \textbf{0.806}\scriptsize{0.127} \\

1.0 & \textbf{0.562}\scriptsize{0.218} & \textbf{0.527}\scriptsize{0.255} & \textbf{0.250}\scriptsize{0.157} & \textbf{0.349}\scriptsize{0.231} & \textbf{0.851}\scriptsize{0.104} & 1.0 & \textbf{0.454}\scriptsize{0.234} & \textbf{0.402}\scriptsize{0.271} & \textbf{0.351}\scriptsize{0.205} & \textbf{0.449}\scriptsize{0.247} & \textbf{0.799}\scriptsize{0.127} \\

1.2 & \textbf{0.541}\scriptsize{0.234} & \textbf{0.505}\scriptsize{0.267} & \textbf{0.263}\scriptsize{0.157} & \textbf{0.366}\scriptsize{0.227} & \textbf{0.842}\scriptsize{0.106} & 1.2 & \textbf{0.470}\scriptsize{0.231} & \textbf{0.428}\scriptsize{0.263} & \textbf{0.337}\scriptsize{0.188} & \textbf{0.432}\scriptsize{0.228} & \textbf{0.805}\scriptsize{0.121} \\

1.4 & \textbf{0.541}\scriptsize{0.236} & \textbf{0.502}\scriptsize{0.272} & \textbf{0.277}\scriptsize{0.165} & \textbf{0.392}\scriptsize{0.240} & \textbf{0.839}\scriptsize{0.105} & 1.4 & \textbf{0.408}\scriptsize{0.220} & \textbf{0.356}\scriptsize{0.255} & \textbf{0.376}\scriptsize{0.177} & \textbf{0.494}\scriptsize{0.233} & \textbf{0.775}\scriptsize{0.121} \\

1.6 & \textbf{0.464}\scriptsize{0.223} & \textbf{0.420}\scriptsize{0.259} & \textbf{0.318}\scriptsize{0.168} & \textbf{0.446}\scriptsize{0.238} & \textbf{0.810}\scriptsize{0.108} & 1.6 & \textbf{0.383}\scriptsize{0.206} & \textbf{0.316}\scriptsize{0.250} & \textbf{0.412}\scriptsize{0.204} & \textbf{0.530}\scriptsize{0.237} & \textbf{0.751}\scriptsize{0.132} \\

\bottomrule
\end{tabular}
\end{table*}

\subsection{Transformation Analysis}

Experimental results are normalised to facilitate analysis and enhance the clarity of visualisations. All quantitative outcomes are scaled from 0 to 1, enabling a clear comparison of the LLM's task selection behaviour. These results illustrate whether the LLM tends toward \textit{perfect alignment} with the to-do list or \textit{no alignment}. For similarity measures such as LCSS, LCP, and SR, a value of 1 indicates perfect alignment with the to-do list, signifying that the LLM follows the original sequence exactly. In contrast, for distance measures like LEV and HAM, a value of 0 represents perfect alignment, meaning no deviations from the original sequence are observed.

\subsubsection{Effect of Sampling Temperature on Sequence Alignment}

Inference hyperparameters such as sampling temperature, top-k sampling, repetition penalty, and maximum token length can be fine-tuned to modify the LLMs output at runtime \cite{touvron2023llama, renze2024effect}. Whilst the focus of this study is centred on the effect of induced personality traits, it remains significant to acknowledge the influence of these hyperparameters with regards to the LLMs performance. Hyperparameters affecting repetition penalties and token lengths are irrelevant in this experiment due to its programmatic design. However, sampling temperature and top-p (nucleus sampling) are directly relevant as these are hyperparameters used to control the randomness and creativity of generated text \cite{openai2023gpt, renze2024effect}. For GPT-based models, it is advised to only adjust one of these parameters, not both \cite{openai2023gpt}. Adjustment of sampling temperature is therefore opted for here.

To empirically investigate the influence adjusted sampling temperature has upon the performance of a GPT-based LLM in the context of our experiment, we perform the experiment outlined in Section \ref{Methodology}, albeit with a strict focus on the control condition–where no personality is induced. The purpose of this evaluation is to test the working hypothesis that increased temperature settings within GPT-based models result in greater levels of non-determinism \cite{ouyang2023llm, openai2023gpt} whilst simultaneously determining the robustness of our measures. By systematically adjusting the sampling temperature, we can observe whether the LLMs adherence to the original to-do list diminishes as randomness increases, thereby validating the effectiveness of our chosen metrics in capturing alignment with the task sequence.

\begin{table*}[ht]
\setlength{\tabcolsep}{7.5pt}
\centering
\small

\caption{Normalised GPT-4o and GPT-4o-Mini results per metric with $\tau = 1.0$ across all groups (OCEAN $\rightarrow$ Low/High).}
\vspace{-5pt}

\label{tab:4o-4o-mini-Combined}
\begin{tabular}{@{}c c c c c c c | c c c c c c c@{}}
\toprule

\multirow{2}{*}{$T$} 
& \multirow{2}{*}{\textit{Dir.}} & \multicolumn{5}{c}{\textbf{GPT-4o} ($\mu$ / $\sigma$)} 

& \multirow{2}{*}{$T$}
& \multirow{2}{*}{\textit{Dir.}} & \multicolumn{5}{c}{\textbf{GPT-4o-Mini} ($\mu$ / $\sigma$)} \\

\cmidrule(r){3-7} \cmidrule(r){10-14}

& & \multicolumn{1}{c}{\textbf{LCSS}} & \multicolumn{1}{c}{\textbf{LCP}} & \multicolumn{1}{c}{\textbf{LEV}} & \multicolumn{1}{c}{\textbf{SR}} & \multicolumn{1}{c}{\textbf{HAM}} 

& & & \multicolumn{1}{c}{\textbf{LCSS}} & \multicolumn{1}{c}{\textbf{LCP}} & \multicolumn{1}{c}{\textbf{LEV}} & \multicolumn{1}{c}{\textbf{SR}} & \multicolumn{1}{c}{\textbf{HAM}} \\
\midrule

\multirow{2}{*}{\centering \textbf{O}} & High & \textbf{0.12}\scriptsize{0.05} & \textbf{0.03}\scriptsize{0.04} & \textbf{0.79}\scriptsize{0.10} & \textbf{0.87}\scriptsize{0.09} & \textbf{0.46}\scriptsize{0.09} & \multirow{2}{*}{\centering \textbf{O}} & High & \textbf{0.14}\scriptsize{0.05} & \textbf{0.05}\scriptsize{0.06} & \textbf{0.72}\scriptsize{0.11} & \textbf{0.86}\scriptsize{0.10} & \textbf{0.53}\scriptsize{0.08} \\

& Low & \textbf{0.19}\scriptsize{0.10} & \textbf{0.16}\scriptsize{0.12} & \textbf{0.65}\scriptsize{0.12} & \textbf{0.74}\scriptsize{0.12} & \textbf{0.55}\scriptsize{0.10} &  & Low & \textbf{0.20}\scriptsize{0.11} & \textbf{0.16}\scriptsize{0.13} & \textbf{0.62}\scriptsize{0.12} & \textbf{0.73}\scriptsize{0.13} & \textbf{0.59}\scriptsize{0.10} \\
\midrule

\multirow{2}{*}{\centering \textbf{C}} & High & \textbf{0.22}\scriptsize{0.12} & \textbf{0.18}\scriptsize{0.14} & \textbf{0.60}\scriptsize{0.15} & \textbf{0.70}\scriptsize{0.15} & \textbf{0.61}\scriptsize{0.11} & \multirow{2}{*}{\centering \textbf{C}} & High & \textbf{0.28}\scriptsize{0.17} & \textbf{0.24}\scriptsize{0.18} & \textbf{0.48}\scriptsize{0.18} & \textbf{0.61}\scriptsize{0.20} & \textbf{0.71}\scriptsize{0.12} \\

& Low & \textbf{0.10}\scriptsize{0.03} & \textbf{0.00}\scriptsize{0.01} & \textbf{0.89}\scriptsize{0.07} & \textbf{0.93}\scriptsize{0.07} & \textbf{0.32}\scriptsize{0.06} &  & Low & \textbf{0.10}\scriptsize{0.04} & \textbf{0.00}\scriptsize{0.01} & \textbf{0.88}\scriptsize{0.07} & \textbf{0.94}\scriptsize{0.06} & \textbf{0.38}\scriptsize{0.07} \\
\midrule

\multirow{2}{*}{\centering \textbf{E}} & High & \textbf{0.12}\scriptsize{0.06} & \textbf{0.03}\scriptsize{0.07} & \textbf{0.79}\scriptsize{0.10} & \textbf{0.87}\scriptsize{0.09} & \textbf{0.45}\scriptsize{0.08} & \multirow{2}{*}{\centering \textbf{E}} & High & \textbf{0.14}\scriptsize{0.05} & \textbf{0.03}\scriptsize{0.04} & \textbf{0.74}\scriptsize{0.09} & \textbf{0.88}\scriptsize{0.09} & \textbf{0.51}\scriptsize{0.07} \\

& Low & \textbf{0.12}\scriptsize{0.05} & \textbf{0.05}\scriptsize{0.06} & \textbf{0.81}\scriptsize{0.09} & \textbf{0.87}\scriptsize{0.10} & \textbf{0.43}\scriptsize{0.07} &  & Low & \textbf{0.12}\scriptsize{0.04} & \textbf{0.07}\scriptsize{0.05} & \textbf{0.82}\scriptsize{0.08} & \textbf{0.88}\scriptsize{0.08} & \textbf{0.44}\scriptsize{0.08} \\
\midrule

\multirow{2}{*}{\centering \textbf{A}} & High & \textbf{0.20}\scriptsize{0.10} & \textbf{0.17}\scriptsize{0.13} & \textbf{0.62}\scriptsize{0.11} & \textbf{0.71}\scriptsize{0.12} & \textbf{0.59}\scriptsize{0.09} & \multirow{2}{*}{\centering \textbf{A}} & High & \textbf{0.19}\scriptsize{0.08} & \textbf{0.14}\scriptsize{0.10} & \textbf{0.62}\scriptsize{0.13} & \textbf{0.76}\scriptsize{0.13} & \textbf{0.61}\scriptsize{0.09} \\

& Low & \textbf{0.17}\scriptsize{0.08} & \textbf{0.11}\scriptsize{0.10} & \textbf{0.66}\scriptsize{0.12} & \textbf{0.78}\scriptsize{0.11} & \textbf{0.56}\scriptsize{0.10} &  & Low & \textbf{0.15}\scriptsize{0.05} & \textbf{0.07}\scriptsize{0.07} & \textbf{0.68}\scriptsize{0.11} & \textbf{0.82}\scriptsize{0.11} & \textbf{0.57}\scriptsize{0.08} \\
\midrule

\multirow{2}{*}{\centering \textbf{N}} & High & \textbf{0.22}\scriptsize{0.10} & \textbf{0.16}\scriptsize{0.13} & \textbf{0.61}\scriptsize{0.12} & \textbf{0.73}\scriptsize{0.14} & \textbf{0.59}\scriptsize{0.10} & \multirow{2}{*}{\centering \textbf{N}} & High & \textbf{0.18}\scriptsize{0.08} & \textbf{0.14}\scriptsize{0.10} & \textbf{0.61}\scriptsize{0.12} & \textbf{0.73}\scriptsize{0.13} & \textbf{0.60}\scriptsize{0.09} \\

& Low & \textbf{0.36}\scriptsize{0.16} & \textbf{0.33}\scriptsize{0.19} & \textbf{0.46}\scriptsize{0.14} & \textbf{0.57}\scriptsize{0.17} & \textbf{0.71}\scriptsize{0.10} &  & Low & \textbf{0.30}\scriptsize{0.16} & \textbf{0.27}\scriptsize{0.18} & \textbf{0.48}\scriptsize{0.15} & \textbf{0.58}\scriptsize{0.17} & \textbf{0.70}\scriptsize{0.11} \\
\midrule

\textbf{B} & N/A & \textbf{0.56}\scriptsize{0.22} & \textbf{0.53}\scriptsize{0.26} & \textbf{0.25}\scriptsize{0.16} & \textbf{0.35}\scriptsize{0.23} & \textbf{0.85}\scriptsize{0.10} & \textbf{B} & N/A & \textbf{0.45}\scriptsize{0.23} & \textbf{0.40}\scriptsize{0.27} & \textbf{0.35}\scriptsize{0.20} & \textbf{0.45}\scriptsize{0.25} & \textbf{0.80}\scriptsize{0.13} \\

\bottomrule
\end{tabular}
\end{table*}

The results, presented in Table \ref{tab:4o-4o-mini-temperature}, demonstrate a clear inverse relationship between temperature settings and the LLMs' adherence to the original to-do list across both GPT-4o and GPT-4o-Mini models. Notably, GPT-4o consistently starts with higher alignment metrics, such as LCSS (0.652) and LCP (0.615) at T0.0, compared to GPT-4o-Mini, which begins at 0.528 and 0.489, respectively. These initial differences suggest that GPT-4o is inherently more capable of maintaining task sequences, likely due to its larger scale or more robust sequence retention abilities \cite{openai2023gpt}. As temperature increases from 0.0 to 1.6, both models show a consistent decrease in LCSS, LCP, and SR, reflecting reduced similarity with the original task order. Conversely, the Levenshtein Distance (LEV) increases with higher temperatures in both models, indicating a greater number of edits needed to align the completed tasks with the to-do list. Interestingly, Hamming Distance (HAM) shows a slight decrease in both models, suggesting that while overall task order becomes more randomised, specific tasks may still align by chance at higher temperatures. Despite these nuances, the findings robustly support the hypothesis that higher temperatures introduce greater variability and less determinism in the LLMs' task selection behaviour, with GPT-4o-Mini demonstrating a baseline tendency towards less deterministic outputs even at lower temperatures.

\subsubsection{Transformation Significance: Experimental Conditions vs. Control} \label{Statistical Significance: Exp. vs Control}

\begin{figure*}[htb]
    \centering
    \includegraphics[width=\linewidth]{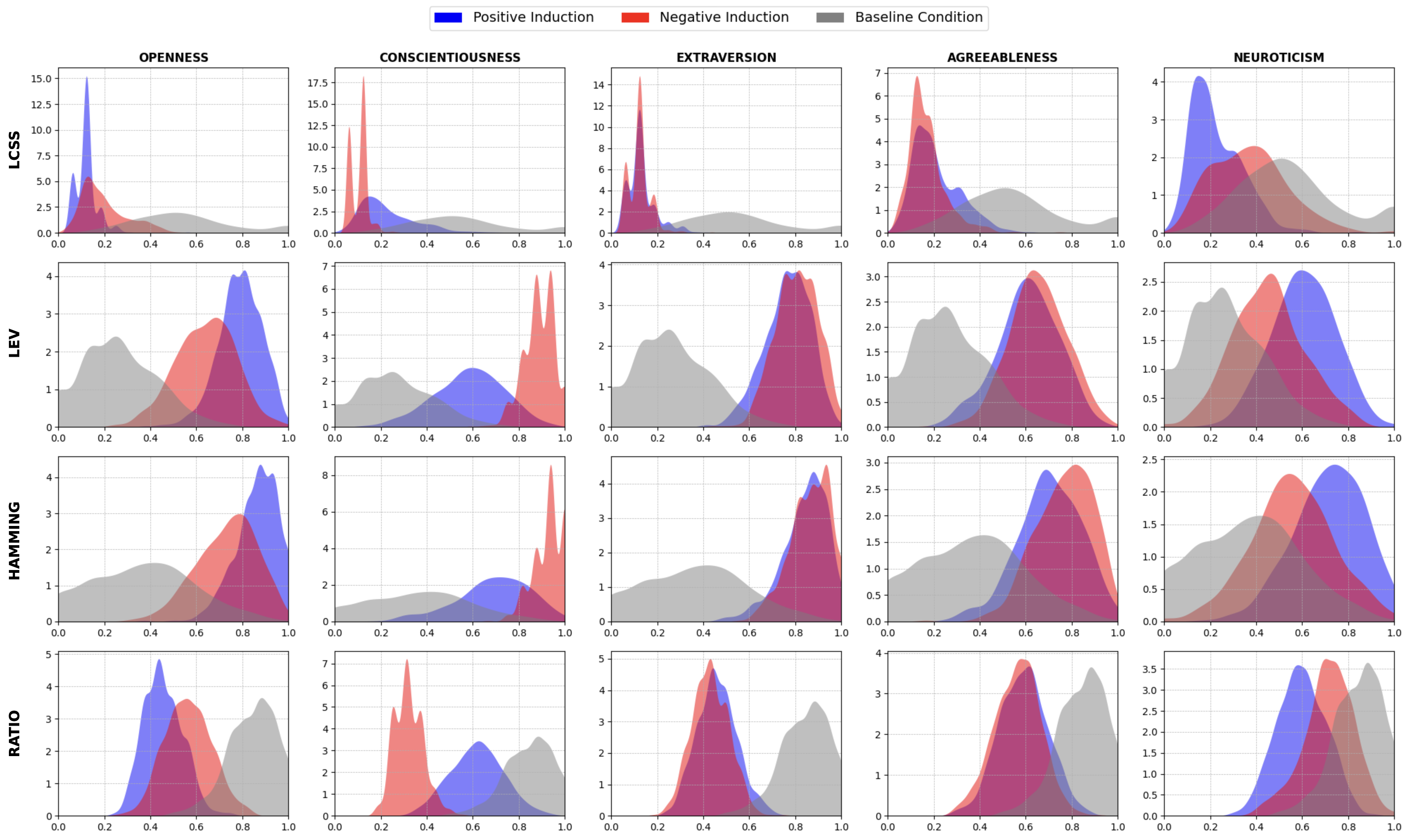}
    \caption{Kernel density estimation plots for GPT-4o across all experimental groups (OCEAN $\rightarrow$ Low/High) per quantitative measure, including the control. Each sub-plot visually represents the GPT-4o results provided in Table \ref{tab:4o-4o-mini-Combined}.}
    \label{fig:4o_All_Results}
    \Description{This image presents the quantitative measures of LCSS, LEV, HAMMING, and RATIO for the GPT-4o model. Represented as a kernel density estimation, it plots together, as independent sub-plots, the computed quantitative values deriving from these metrics, per each trait, on the conditions of: positive induction, negative induction, and baseline condition (no personality induction).}
\end{figure*}

The sequence difference (or similarity) metrics were also applied to assess whether the induction of a personality has a material impact on the populations of the measurements of the selected metrics. For this analysis, the LLMs GPT-4o, GPT-4o-mini and GPT-3.5-Turbo were assessed using the experimental framework outlined previously. Figure \ref{fig:4o_All_Results} illustrates the corresponding distributions for GPT-4o per each of the key metrics (LCSS, LEV, HAMMING, RATIO) only. GPT-4o-Mini produced similar outputs to GPT-4o which is expected given it is a more compact model, sacrificing some performance for greater accessibility and affordability. Thus, visualised results for GPT-4o-Mini were excluded. In relation to GPT-3.5-Turbo, a predecessor model, the observed results across all metrics did not differ substantially, despite being found to be statistically significant in most cases (See \ref{GPT-3.5-Turbo}).

In this analysis we seek to assess whether the induction of a persona creates a statistically significant different population of the metrics when compared to the baseline. The Null hypothesis states there will be no difference to the baseline population with the Alternative Hypothesis being that the induced personality trait will produce a different population of measurements. To test this, an independent two-sample Welch's t-test is performed between each experimental condition and the control for each measure. For a singular model (\textit{i.e.}, GPT-4o), this results in 50 pairwise comparisons (5 traits x 2 directions x 5 measures). For example, "LCSS for Openness (High)" is compared against "LCSS for Baseline". To control for the family-wise error rate (FWER) due to multiple comparisons, the Bonferroni correction is applied, controlled by the following:
% \begin{small}
\[
\scriptsize
\text{FWER} = P\left( \bigcup_{i=1}^{m_0} \left\{ p_i \leq \frac{\alpha}{m} \right\} \right) \leq \sum_{i=1}^{m_0} P\left( p_i \leq \frac{\alpha}{m} \right) = m_0 \frac{\alpha}{m} \leq \alpha
\]
% \end{small}

where $p_i$ is the p-value for the $i$-th test, $m = 50$ is the total number of tests, and $\alpha = 0.05$ is the overall significance level. With 50 tests, the alpha level ($p \leq 0.05$) is then adjusted to $p \leq 0.001$.

\subsubsection{GPT-4o and GPT-4o-Mini} \label{GPT-4o and GPT-4o-Mini}

For GPT-4o, the Null hypothesis is rejected on all experimental conditions, potentially attributed toward the significantly different results for the baseline group on all measures. For GPT-4o-Mini, the Null hypothesis is rejected on all experimental conditions. Similar to GPT-4o, this may be attributed toward the significantly different results for the baseline group on all measures. All results are displayed in Table \ref{tab:4o-4o-mini-Combined}.

\subsubsection{GPT-3.5-Turbo} \label{GPT-3.5-Turbo}

Null hypothesis is rejected on 41 experimental conditions, accepted on 9. Statistically insignificant results, displayed in Table \ref{tab:gpt3.5-tabular-results}, are not emboldened and marked with an asterisk (*). Whilst the statistical test results indicate a significantly different task ordering for the majority of experimental conditions and quantitative measures, the absolute difference, between all experimental conditions and that of the control, are far less considerable when compared to GPT-4o and GPT-4o-Mini (Table \ref{tab:4o-4o-mini-Combined}).

\begin{table}[ht]
\centering
\small
\captionsetup{font=small}

\caption{Normalised GPT-3.5-Turbo results per metric with $\tau = 1.0$ across all experimental groups (OCEAN $\rightarrow$ Low/High). B = Control\footnotesize\textsuperscript{1}.}
\vspace{-5pt}

\label{tab:gpt3.5-tabular-results}
\begin{tabular}{@{}cc cc c c c@{}}
\toprule
\multirow{2}{*}{\centering} & \multirow{2}{*}{\centering} 
& \multicolumn{5}{c}{\textbf{GPT-3.5-Turbo} ($\mu$ / $\sigma$)} \\
\cmidrule(r){3-7}
 &  & \multicolumn{1}{c}{\textbf{LCSS}} & \multicolumn{1}{c}{\textbf{LCP}} & \multicolumn{1}{c}{\textbf{LEV}} & \multicolumn{1}{c}{\textbf{SR}} & \multicolumn{1}{c}{\textbf{HAM}} \\
\midrule

\multirow{2}{*}{\centering \textbf{O}} & High & \textbf{0.111}\scriptsize{0.04} & 0.001*\scriptsize{0.006} & \textbf{0.859}\scriptsize{0.082} & \textbf{0.93}\scriptsize{0.064} & \textbf{0.404}\scriptsize{0.077} \\

& Low & 0.116*\scriptsize{0.048} & \textbf{0.008}\scriptsize{0.023} & \textbf{0.817}\scriptsize{0.089} & \textbf{0.913}\scriptsize{0.07} & \textbf{0.447}\scriptsize{0.077} \\
\midrule

\multirow{2}{*}{\centering \textbf{C}} & High & 0.116*\scriptsize{0.042} & \textbf{0.002}\scriptsize{0.011} & \textbf{0.823}\scriptsize{0.091} & \textbf{0.91}\scriptsize{0.072} & \textbf{0.432}\scriptsize{0.076} \\

& Low & \textbf{0.112}\scriptsize{0.042} & \textbf{0.002}\scriptsize{0.011} & \textbf{0.841}\scriptsize{0.085} & \textbf{0.915}\scriptsize{0.073} & \textbf{0.41}\scriptsize{0.076} \\
\midrule

\multirow{2}{*}{\centering \textbf{E}} & High & \textbf{0.115}\scriptsize{0.039} & \textbf{0.001}\scriptsize{0.007} & \textbf{0.83}\scriptsize{0.089} & \textbf{0.916}\scriptsize{0.07} & \textbf{0.432}\scriptsize{0.079} \\

& Low & \textbf{0.115}\scriptsize{0.041} & 0.006*\scriptsize{0.019} & \textbf{0.855}\scriptsize{0.083} & \textbf{0.925}\scriptsize{0.07} & \textbf{0.416}\scriptsize{0.072} \\
\midrule

\multirow{2}{*}{\centering \textbf{A}} & High & 0.117*\scriptsize{0.04} & \textbf{0.004}\scriptsize{0.016} & \textbf{0.822}\scriptsize{0.09} & \textbf{0.914}\scriptsize{0.074} & \textbf{0.438}\scriptsize{0.076} \\

& Low & \textbf{0.114}\scriptsize{0.043} & \textbf{0.003}\scriptsize{0.012} & \textbf{0.843}\scriptsize{0.084} & \textbf{0.928}\scriptsize{0.063} & \textbf{0.42}\scriptsize{0.078} \\
\midrule

\multirow{2}{*}{\centering \textbf{N}} & High & \textbf{0.112}\scriptsize{0.041} & 0.005*\scriptsize{0.02} & \textbf{0.839}\scriptsize{0.085} & \textbf{0.92}\scriptsize{0.072} & \textbf{0.425}\scriptsize{0.072} \\

& Low & 0.119*\scriptsize{0.047} & 0.005*\scriptsize{0.018} & \textbf{0.802}\scriptsize{0.098} & 0.913*\scriptsize{0.078} & \textbf{0.458}\scriptsize{0.088} \\
\midrule

\textbf{ B}\textsuperscript{1} & N/A & \textbf{0.124}\scriptsize{0.044} & \textbf{0.008}\scriptsize{0.025} & \textbf{0.776}\scriptsize{0.093} & \textbf{0.892}\scriptsize{0.082} & \textbf{0.472}\scriptsize{0.081} \\

\bottomrule
\end{tabular}
\end{table}
\section{Conclusion} \label{Conclusion}

% AAMAS 2025 encourages you to carefully consider the potential impact of your research on society, and to discuss any significant ethical, societal, or legal concerns that may arise.

This study proposed a method to quantitatively measure the effect of prompt-based personality induction on LLM-based agent decision-making in task selection, scheduling, and planning. While the display of synthetic personality in LLMs is known, no studies have empirically focused on evaluating how inducing personality traits based on the Five-Factor OCEAN model impacts critical agent behaviours governing decision-making, particularly task selection.

Experiments with all OCEAN traits and a neutral control group showed that personality induction leads to significant differences in task prioritisation across GPT-4o, GPT-4o-Mini, and GPT-3.5-Turbo models. Effects were more pronounced in GPT-4o models, indicating their greater capacity for reasoning and exhibiting these traits. Analysis specifically examined the \textit{degree of plausibility} in how the LLM-based agent prioritised tasks, assessing whether the agent’s decisions aligned with contemporary psychological understandings of each trait. Our results showed that traits like \textit{Openness}, \textit{Conscientiousness}, and \textit{Extraversion} substantially impacted task prioritisation, causing significant deviations from original schedules. In contrast, \textit{Agreeableness} and \textit{Neuroticism} had less pronounced effects, possibly due to LLMs being less receptive to these traits or due to limitations in task relevance.

These findings show that large-scale, pre-trained language models like GPT-4o can exhibit personality traits, confirming previous studies and demonstrating impact within a downstream task. Moreover, our validated personality induction method highlights potential for enhancing autonomous agents in planning and scheduling, especially in systems performing tasks akin to humans. Our contributions lay a foundation for integrating nuanced human-like behaviours into autonomous systems, enhancing their effectiveness in environments requiring sophisticated, plausible decision-making.

\subsection{Ethical Considerations}

This research aims to enhance the development of Deceptive Agents, a novel class of LLM-based autonomous agents which are intended to effectuate highly plausible simulacra of humans interacting with digital systems for cyber defense through strategic deception \cite{newsham2024sandman}. Ethically, this approach operates on the principle of `rightful deception', where targets have no legitimate claim to truth or transparency due to their unethical intent to access or damage systems without authorisation. Whilst the intended use is strictly within a defensive context, there exists the possibility that this work could facilitate the generation of misinformation or enable tailored persuasion tactics, thereby exacerbating issues related to manipulation and deceit in broader contexts \cite{park2023generative}. Specifically, the ability to induce personality traits within LLMs could be exploited to create more convincing deceptive content, potentially undermining public trust. It is therefore essential to implement safeguards and adhere to ethical guidelines in any research or application involving the induction of personality traits within LLMs to mimic human values, thought patterns, and behaviour. This includes ensuring robust oversight, bias mitigation, and alignment to prevent misuse.

%%%%%%%%%%%%%%%%%%%%%%%%%%%%%%%%%%%%%%%%%%%%%%%%%%%%%%%%%%%%%%%%%%%%%%%%

%%% The acknowledgments section is defined using the "acks" environment
%%% (rather than an unnumbered section). The use of this environment 
%%% ensures the proper identification of the section in the article 
%%% metadata as well as the consistent spelling of the heading.

\begin{acks}
This research is supported by FUJITSU Enterprise \& Cyber Security, to whom we extend our gratitude for their funding and support.
\end{acks}

%%%%%%%%%%%%%%%%%%%%%%%%%%%%%%%%%%%%%%%%%%%%%%%%%%%%%%%%%%%%%%%%%%%%%%%%

%%% The next two lines define, first, the bibliography style to be 
%%% applied, and, second, the bibliography file to be used.

\bibliographystyle{ACM-Reference-Format} 
\bibliography{bibliography}

%%%%%%%%%%%%%%%%%%%%%%%%%%%%%%%%%%%%%%%%%%%%%%%%%%%%%%%%%%%%%%%%%%%%%%%%

\end{document}